\def\year{2022}\relax
\documentclass[letterpaper]{article} 
\usepackage{aaai22}  
\usepackage{times}  
\usepackage{helvet}  
\usepackage{courier}  
\usepackage[hyphens]{url}  
\usepackage{graphicx} 
\usepackage{natbib}  
\usepackage{caption} 
\DeclareCaptionStyle{ruled}{labelfont=normalfont,labelsep=colon,strut=off} 
\usepackage{algorithm}
\usepackage{algorithmic} 

\usepackage{amsmath}

\usepackage{newfloat}
\usepackage{listings}
\floatstyle{ruled}
\newfloat{listing}{tb}{lst}{}
\floatname{listing}{Listing}

\title{Predicting blood pressure under circumstances of  missing data: An analysis of missing data patterns and imputation methods using NHANES}

\author {
     Harish Chauhan$^*$,
     Nikunj Gupta$^*$,
     Zoe Haskell-Craig$^*$
 }
\affiliations {
    New York University\\
}

\begin{document}

\maketitle
\def\thefootnote{*}\footnotetext{These authors contributed equally to this work.}\def\thefootnote{\arabic{footnote}}



\section{Motivation} 

The World Health Organization defines cardio-vascular disease (CVD) as “a group of disorders of the heart and blood vessels,” including coronary heart disease and stroke. \cite{c:22} CVD is affected by “intermediate risk factors” such as raised blood pressure, raised blood glucose, raised blood lipids, and obesity. These are predominantly influenced by lifestyle and behaviour, including physical inactivity, unhealthy diets, high intake of salt, and tobacco and alcohol use. However, genetics and social/environmental factors such as poverty, stress, and racism also play an important role. Researchers studying the behavioural and environmental factors associated with these “intermediate risk factors” need access to high quality and detailed information on diet and physical activity. 
 
However, missing data are a pervasive problem in clinical and public health research, affecting both randomized trials and observational studies. Reasons for missing data can vary substantially across studies because of loss to follow-up, missed study visits, refusal to answer survey questions, or an unrecorded measurement during an office visit. One method of handling missing values is to simply delete observations for which there is missingness (called Complete Case Analysis). This is rarely used as deleting the data point containing missing data (List wise deletion) results in a smaller number of samples and thus affects accuracy. Additional methods of handling missing data exists, such as summarizing the variables with its observed values (Available Case Analysis).

Motivated by the pervasiveness of missing data in the NHANES dataset, we will conduct an analysis of imputation methods under different  simulated patterns of missing data. We will then apply these imputation methods to create a complete dataset upon which we can use ordinary least squares to predict blood pressure from diet and physical activity.

\section{Methodology} 

\subsection{Data}
We used data from the 2013-2014 National Health and Nutrition Examination Survey (NHANES), a study that examined a nationally representative sample of the US population. Health interviews and survey questionnaires were administered to participants of all ages, alongside laboratory tests and a physical examination. \cite{c:23} Data collected included demographic data (age, gender, race and ethnicity, education level, citizenship status, household size and income), measures of body proportions and weight, pulse and blood pressure, history of diabetes, dietary practices (such as use of table salt), measure of caloric and macro-nutrient intake (dietary fiber, fat, etc), type and duration of physical activity, and blood cholesterol levels. 

The final sample size for 2013-2014 was 10 175. After limiting the sample to participants aged 20 or older (as some tests were only performed on adults), we had n = 5769 observations. After limiting participants to those without missing data for day 1 systolic blood pressure, we had 5111 rows. High low-density lipoproteins (LDL) cholesterol is considered especially dangerous for CVD as it can build up in blood vessels causing blockages. We categorized participants as having high LDL if they had more than 160 mg/dL (milligrams per deciliter) in their blood. \cite{c:24}

Exploratory data analysis was conducted by examining pairwise Pearson's correlation coefficients between variables. Additionally, patterns of missingness were plotted. 

\subsection{Missing Data Mechanisms}
Restricting our analysis to 16 variables including diet column DR1TKCAL (table \ref{tab:vars}), and participants for whom we observe data for each variable (no missing data), we simulated missing data according to missing completely at random (MCAR), missing at random (MAR), and missing not at random (MNAR) patterns.

\begin{table} [!htb]
\begin{center}
\begin{tabular}{| c | c |} 
\hline 
 Variable Name & Description \\
 \hline \hline
 RIDAGEYR & Age  \\  
 \hline
 RIAGENDR & Gender \\ 
 \hline 
 RIDRETH1[3] &  Race/Ethnicity \\ 
 \hline
 DMDCITZN & Citizenship  \\ 
 \hline
 BMXLEG & Leg circumference  \\ 
 \hline
 BPXPULS & Pulse  \\ 
 \hline
 DIQ010 & Diabetes diagnosis  \\ 
 \hline
 DIQ050 & Insulin use \\ 
 \hline
 HIQ011 & Health insurance \\
 \hline
 PAQ635 & Walking or bicycling \\
 \hline
 PAQ650 & Vigorous exercise \\
 \hline
 PAQ665 & Moderate exercise\\
 \hline
 PAD680 & Sedentary minutes\\
 \hline
 PAQ710 & TV use\\
 \hline
 DR1TKCAL & Caloric intake\\
 \hline
\end{tabular}
\caption{Variables included in data subset used for missing data analysis.}
\label{tab:vars}
\end{center}
\end{table}

\paragraph{Missingness completely at random (MCAR)} A variable is missing completely at random if the probability of missingness is the same for all units, that is, the missingness is independent of all observed and unobserved variables. \cite{c:83} If data are missing completely at random, then throwing out cases with missing data does not bias your inferences. We simulated MCAR by randomly removing observations from our sample.

\paragraph{Missingness at random (MAR)} Most missingness is not completely at random, as can be seen from the data themselves. A more general assumption, missing at random, is that the probability a variable is missing depends only on available information. That is, the missingness is related to the observed but not the unobserved variables. \cite{c:83} Thus, if sex, age, bmi, and diet are recorded for all the people in the survey, then caloric intake is missing at random if the probability of nonresponse to this question depends only on these other, fully recorded variables. We simulated MAR data by removing data from the diet column DR1TKCAL at twice the probability if the respondent had a value of 2 for the physical activity question.

\paragraph{Missingness not at random (MNAR)} Missingness is no longer “at random” if it depends on information that has not been recorded and this information also predicts the missing values. For example, suppose that “surly” people are less likely to respond to the ``dietary" question, surliness is predictive of blood pressure, and “surliness” is unobserved. Then, the blood pressure values are not missing at random. A familiar example from medical studies is that if a particular treatment causes discomfort, a patient is more likely to drop out of the study. This missingness is not at random (unless “discomfort” is measured and observed for all patients). If missingness is not at random, it must be explicitly modeled, or else you must accept some bias in your inferences. We simulated MNAR by removing data from the diet column at twice the probability if the respondent had a BMI greater than 25 - the definition of overweight. Note that while BMI is present in the larger dataset, we did not include this in our sample for missingness simulation - thus it takes on the role of an unobserved variable.

\subsection{Imputation methods} 
We selected following imputation methods for our analysis.\\ \\
\textbf{Univariate Feature Imputation }method employs data imputation either by mean, median or most frequent values of a column in which missing data is located. This strategy can severely distort the distribution for this variable, leading to complications with summary measures including, notably, underestimates of the standard deviation. Moreover, mean imputation distorts relationships between variables by “pulling” estimates of the correlation toward zero.\\ \\
\textbf{Nearest Neighbors Imputation} provides imputation by finding the samples in the set “closest” to it and averages these nearby points to fill in the value. It is certainly advantageous than univariate feature imputation but this method is computationally heavy and it is quite sensitive to outliers.\\ \\
\textbf{Low Rank Approximation} is approximating a matrix by one whose rank is less than that of the original matrix using Singular Value Decomposition (SVD). 
This technique iteratively approximates the missing observations from the complete low-rank matrix then recomputes the SVD. We created the SVD on a training dataset and then chose the rank to minimize error in the test set. \\ \\
\textbf{Multivariate Feature Imputation }method is a more sophisticated approach. It models each feature with missing values as a function of other features, and uses that estimate for imputation. It does so in an iterated round-robin fashion: at each step, a feature column is designated as output Y and the other feature columns are treated as inputs X. A regressor is fit on (X,Y) for known Y. Then, the regressor is used to predict the missing values of Y. This is done for each feature in an iterative fashion.\\ \\
\textbf{Evaluation: Mean Squared Error}
We used the mean squared error (MSE) to evaluate the performance of the imputation method, comparing the imputed values to those in the original subsample.
\subsection{Prediction}
\textbf{OLS Regression On Imputed Data} 
In order to understand the association between lifestyle and blood pressure, an important "intermediate" risk factor for CVD, we applied Ordinary Least Squares (OLS) regression in tandem with the imputation methods to our complete dataset. OLS is a type of linear least squares method for estimating the unknown parameters in a linear regression model. The resulting estimator can be expressed by a simple formula, especially in the case of a simple linear regression, in which there is a single regressor on the right side of the regression equation. When the data contains many more observations than features (as in our case), OLS regression is optimal in the class of linear unbiased estimators when the errors are homoscedastic and serially uncorrelated. OLS provides minimum-variance mean-unbiased estimation when the errors have finite variances. Under these conditions OLS significantly outperforms lasso and ridge regression (with $\lambda > 0$). 

We first applied the six imputation methods to create a complete dataset without any missing observations. We then split the data 70/30 into training and test sets. We trained the OLS model on the training set, and then applied the model to the test set data to predict blood pressure. We assessed the performance of the various imputation methods using mean squared error (MSE) and root mean squared error (RMSE). 

\section{Results and Discussion} 

\begin{figure} [!htb]
\centering
\includegraphics[width=.6\linewidth]{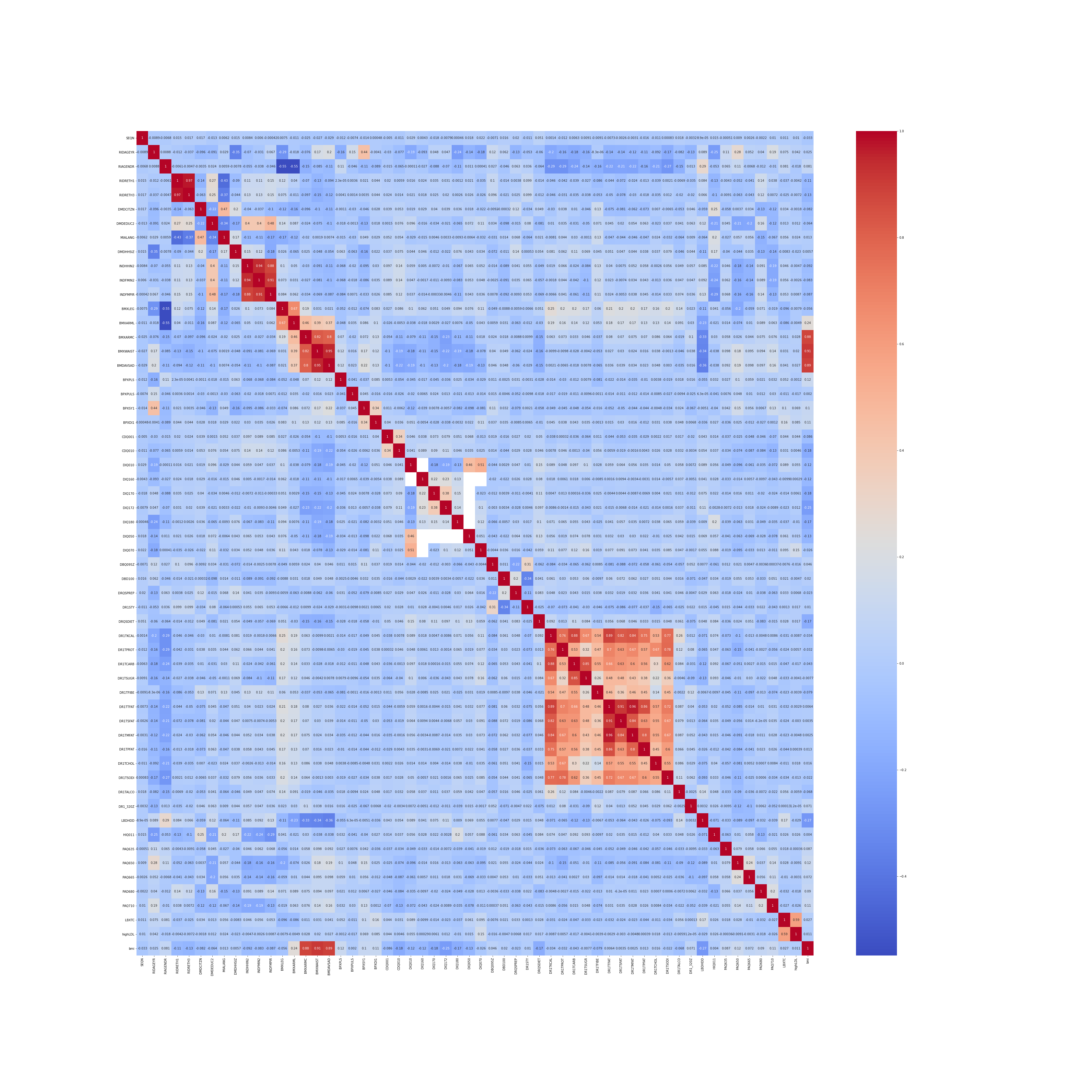}
\caption{Correlation between features}
\label{fig:corr}
\end{figure}

\begin{figure} [!htb]
\centering
\includegraphics[width=.8\linewidth]{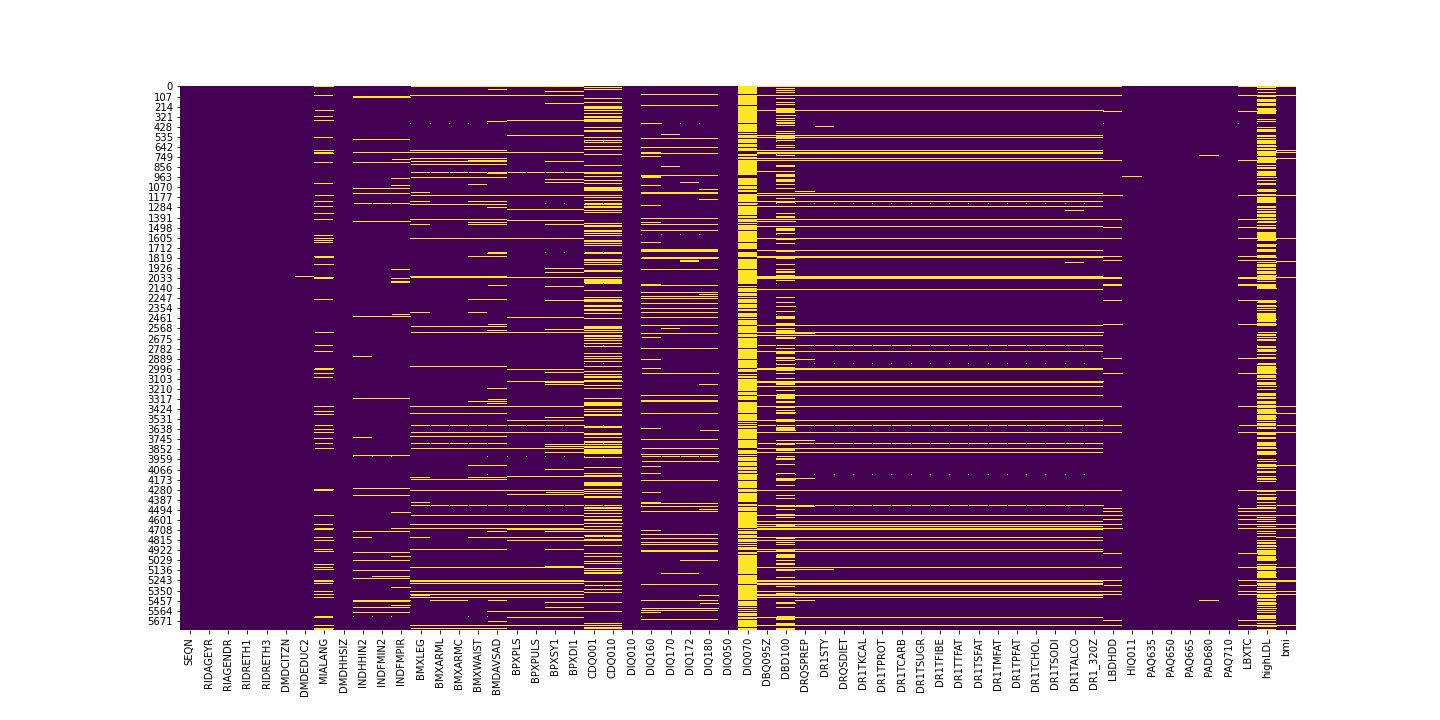}
\caption{Missing data in NHANES dataset}
\label{fig:miss}
\end{figure}

\textbf{Exploratory Data Analysis:}
Correlation plots showed moderate to low correlation between most variables, with the exception of the dietary intake variables, diastolic and systolic blood pressure measured on two separate occasions, and measures of leg, arm, and waist circumference (figure \ref{fig:corr}). Dietary variable missingness occurred concurrently, as did history of cardiovascular disease questionnaire missingness. Using a chi-squared test, we determined that missing dietary data was independent of health insurance status and gender. Missing dietary data was not independent of physical exercise status (variable PAQ650)(figure \ref{fig:miss}).

\textbf{Mean squared error of imputation method by missingness pattern:} 
We considered three general “missingness mechanisms:” MCAR, MAR, and MNAR. We calculate the mean squared error (MSE) between the imputed missing values and the corresponding true values. \vspace{1mm}

\begin {table}[H]
\begin{center}
\begin{tabular}{ |p{3cm}||p{1.4cm}|p{1.4cm}|p{1.4cm}|  }
 \hline
 \multicolumn{4}{|c|}{Mean Squared Error} \\
 \hline
 Imputation Method & MCAR & MAR & MNAR\\
 \hline
 Mean               & 0.831    & 1.149       & 1.045 \\
 Median             & 0.910    & 1.262       & 1.070 \\
 Most-frequent      & 1.157    & 1.484       & 1.188 \\
 KNN                & 0.861    & 1.473       & 0.926 \\
 Low Rank Model     & 1.021    & 1.060       & 1.010 \\
 Regression         & \textbf{0.560 }   & \textbf{1.013}       & \textbf{0.853} \\
 \hline
\end{tabular}
\caption {Analysis of Imputation Methods on Simulated Data} \label{tab:mse} 
\end{center}
\end {table}

Table \ref{tab:mse} summaries our findings of performances of various imputation methods on the simulated missing data. Our study demonstrated that multivariate regression feature imputation outperformed all other imputation methods independent of the pattern of missingness. Aside from regression, the second best performance came from the low rank approximation imputation when the data was MAR or MNAR. However, imputation with the mean performed second-best when the data was MCAR, closely competing with KNN's performance. \vspace{2mm}

\begin{table}[H]
\begin{center}
\begin{tabular}{ |p{3cm}||p{1.4cm}|p{1.4cm}|}
 \hline
 \multicolumn{3}{|c|}{Root Mean Squared Error} \\
 \hline
 Imputation Method & Train & Test\\
 \hline
 Mean               & 14.516    & 14.783                \\
 Median             & 14.776    & \textbf{14.227 }      \\
 Most-frequent      & 14.573   & 14.619                 \\
 KNN                & 14.619    & 14.642                \\
 Low Rank Model     & 18.232    & 17.854                \\
 Regression         & \textbf{14.361 }   & 15.088       \\
 \hline
\end{tabular}
\caption {Regressing blood pressure using Imputed datasets} \label{tab:reg} 
\end{center}
\end{table}

\textbf{Prediction Results:} Table \ref{tab:reg} summarizes the Linear Regression results obtained the complete dataset to predict blood pressure, we found that the model performed best when the data was imputed using \textit{multivariate feature imputation}. Surprisingly, imputing with the median column values resulted in the best performing model on the test set. 

\textbf{Summary:} There are several reasons why we may have obtained this surprising result. For one, we used only a small subset of the features and observations to create the dataset for the simulation analysis. The distribution of data and relationship between features may be quite different in the complete dataset. This could result in the different performances of imputation methods between the complete dataset and the subset. Second, the poor performance of regression imputation in the complete dataset could be explained by the limited correlation between features in the complete dataset. If there is no way to predict one variable from another than regression imputation is likely to perform poorly. Finally, using the column median to impute missing values essentially smooths the data. It is possible that the OLS model predicts better when the features are "smoother".

\section{Conclusion}

To conclude, we found that multivariate regression imputation performed the best across all three simulated patterns of missingness in this dataset. We were able to predict blood pressure from diet and physical activity with reasonable RMSE. Imputing missing values using the column median produced a prediction model with the best RMSE.

This study has several strengths. First, we assessed the performance of the imputation methods under different patterns of missingness, instead of simply assuming all the data was MCAR. Second, we examined the impact of different imputation methods both on the accuracy of imputed values and on the ability to predict blood pressure. This study also has several limitations. We only examined a small subset of the features and observations in our simulation study. Additionally, we only created missingness in one column: caloric intake. Further research should simulate different frequencies of missing values and missing patterns in other features.

\newpage

\bibliography{aaai22}

\end{document}